\documentclass{bmvc2k}

\usepackage{times}
\usepackage{epsfig}
\usepackage{graphicx}
\usepackage{multicol}
\usepackage{amsmath}
\usepackage{amssymb}
\usepackage{fancyhdr}
\usepackage[ruled,vlined]{algorithm2e}
\include{pythonlisting}
\usepackage{subfigure}
\usepackage{varwidth}
\usepackage{subfigure}
\usepackage{xcolor}

\newcommand{\stdv}[1]{\scriptsize$\pm$#1}

\newcommand{\xoverline}[2][0.75]{%
    \sbox{\myboxA}{$\m@th#2$}%
    \setbox\myboxB\null
    \ht\myboxB=\ht\myboxA%
    \dp\myboxB=\dp\myboxA%
    \wd\myboxB=#1\wd\myboxA
    \sbox\myboxB{$\m@th\overline{\copy\myboxB}$}
    \setlength\mylenA{\the\wd\myboxA}
    \addtolength\mylenA{-\the\wd\myboxB}%
    \ifdim\wd\myboxB<\wd\myboxA%
       \rlap{\hskip 0.5\mylenA\usebox\myboxB}{\usebox\myboxA}%
    \else
        \hskip -0.5\mylenA\rlap{\usebox\myboxA}{\hskip 0.5\mylenA\usebox\myboxB}%
    \fi}

\title{Improving Face Recognition with Large Age Gaps by Learning to Distinguish Children}

\addauthor{Jungsoo Lee*}{bebeto@kaist.ac.kr}{1}
\addauthor{Jooyeol Yun*}{blizzard072@kaist.ac.kr}{1}
\addauthor{Sunghyun Park}{psh01087@kaist.ac.kr}{1}
\addauthor{Yonggyu Kim}{dydrb2535@kaist.ac.kr}{2}
\addauthor{Jaegul Choo}{jchoo@kaist.ac.kr}{1}

\addinstitution{
 KAIST AI, \\ South Korea
}
\addinstitution{
 Korea University, \\ South Korea
}

\runninghead{Lee et al}{Improving Face Recognition with Large Age Gaps}

\def\eg{\emph{e.g}\bmvaOneDot}

\def\ie{\emph{i.e}\bmvaOneDot, }
\def\etal{\emph{et al}\bmvaOneDot}

\definecolor{MyGreen}{rgb}{0,0.6,0.3}
\definecolor{blackpink}{rgb}{0.6,0,0.6}
\definecolor{blue}{RGB}{0, 0, 200}
\definecolor{red}{RGB}{255, 0, 0}
\definecolor{black}{RGB}{0, 0, 0}

\definecolor{blue_js}{RGB}{0,26,102}

\begin{document}

\maketitle
\vspace{-0.7cm}
\begin{abstract}
Despite the unprecedented improvement of face recognition, existing face recognition models still show considerably low performances in determining whether a pair of child and adult images belong to the same identity.
Previous approaches mainly focused on increasing the similarity between child and adult images of a given identity to overcome the discrepancy of facial appearances due to aging.
However, we observe that reducing the similarity between child images of different identities is crucial for learning distinct features among children and thus improving face recognition performance in child-adult pairs.
Based on this intuition, we propose a novel loss function called the Inter-Prototype loss which minimizes the similarity between child images. 
Unlike the previous studies, the Inter-Prototype loss does not require additional child images or training additional learnable parameters.
Our extensive experiments and in-depth analyses show that our approach outperforms existing baselines in face recognition with child-adult pairs.
Our code and newly-constructed test sets of child-adult pairs are available at this link~\footnote{\url{https://github.com/leebebeto/Inter-Prototype}}.
\end{abstract}

\vspace{-0.7cm}
\section{Introduction}
\label{sec:intro}
\vspace{-0.3cm}
Although recent face recognition models achieve remarkable improvements~\cite{arcface, sphereface, cosface,curricular, MV-softmax, deepface, deepface_cvpr}, they still show low performance in deciding whether child-adult pairs match to the same identity.
Performing face recognition with child-adult pairs requires models to consider the changes of facial appearances of a given identity (\eg added wrinkles or changes of chin shapes) which occurs due to the aging process.
Such issue needs to be resolved when considering its application such as finding missing children which requires face recognition models to match childhood and grown-up images of a given identity.

\begin{figure}[t!]
    \centering
    \includegraphics[width=0.65\textwidth, clip]{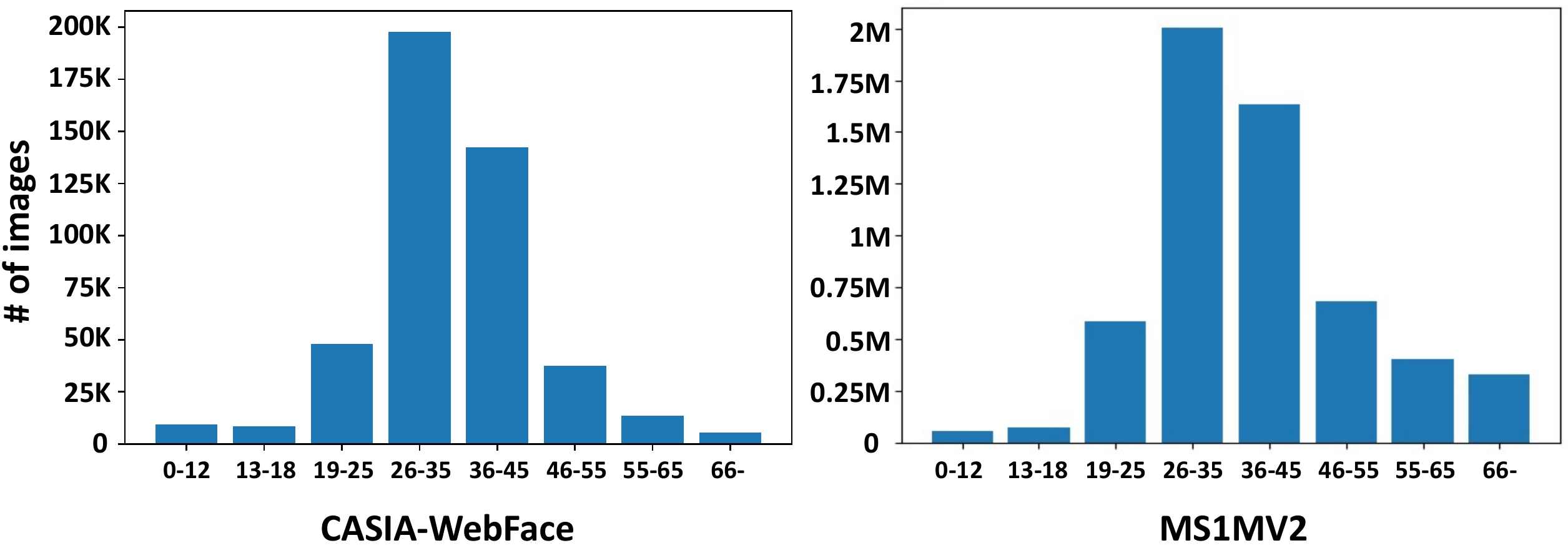}
    \vspace{-0.3cm}
    \caption{Pseudo-labeled age distribution of the CASIA-WebFace dataset~\cite{casia} and MS1MV2. The small number of child images in the dataset limits the model in learning the representation of child images.}
    \vspace{-0.5cm}
    \label{fig:casia_agedist}
\end{figure}

This task remains challenging due to the fact that the child images are excessively scarce in publicly available datasets (\eg CASIA-WebFace~\cite{casia} and MS1MV2~\footnote{MS1MV2 is a refined version of the MS-Celeb-1M dataset.~\cite{ms1mv2}}), as shown in Fig.~\ref{fig:casia_agedist}. 
It requires a non-trivial amount of time to collect both childhood and adulthood images of an identity when building a new dataset.
Even worse, we empirically find that child images generally share appearances such as big eyes, rounded shape of chins, and short hair, which inhibits the face recognition models from extracting the distinct features among child images~\cite{DAL}.
These reasons altogether make it challenging for models to learn the representation of children and verify pairs of child and adult images.~\cite{sexist}.

Recent face recognition models utilize various loss functions~\cite{arcface, cosface, curricular, MV-softmax, broadface} in order to increase the similarity of features extracted from images within an identity (\ie intra-class similarity).
Through an in-depth analysis, we observe that existing loss functions encourage the models to increase the intra-class similarity to a reasonable level.
However, we find that the similarities between different classes (\ie inter-class similarity) among child images are substantially higher than those of adult images, which hinders the model in learning the discriminative features of a child image (Section.~\ref{sec:motivation}).

In this work, we propose a novel loss function called the Inter-Prototype loss which minimizes the inter-class similarity between identities including child images.
With the reduced inter-class similarity, our approach enables the models to learn the discriminative features of a child image, which improves the face recognition performance in child-adult pairs. 
Compared to the previous age-invariant face recognition models~\cite{mtl, DAL, AECNN, OECNN, look-across-elapse} which require a substantial amount of the child images in the training dataset, our approach shows superior performances over existing baselines even with the limited number of child images in the training dataset. Our contributions are as follows:
\vspace{-0.2cm}
\begin{itemize}
    \item We provide an in-depth analysis on existing approaches and reveal that reducing the inter-class similarity between children is crucial for improving face recognition including child-adult pairs.
    \vspace{-0.15cm}
    \item We propose a novel loss function called the Inter-Prototype loss which minimizes the inter-class similarity of the identities including child images in a \emph{simple yet effective} manner.
    \vspace{-0.15cm}
    \item Our approach outperforms the existing baselines on face verification/identification of child-adult pairs with large age gaps. We also release the test sets of child-adult pairs.
\end{itemize}

\vspace{-0.7cm}
\section{Related Work}
\label{sec:related-work}
\vspace{-0.25cm}
Recent advancements of face recognition models~\cite{sphereface,cosface,arcface} follow the line of work where models learn the feature embeddings through training a classifier.
These studies interpret the weight of the last fully-connected layer of a classifier as the representative vectors for each identity in the train set, which we term as \emph{prototype vectors}.
They utilize the margin-based loss functions~\cite{cosface,arcface, sphereface} that assign an additional penalty to the angles between extracted features and their corresponding prototype vectors.
Such margin penalty using the prototype vectors increases the intra-class similarity.
Recent studies~\cite{MV-softmax, curricular} further improve the margin-based loss functions by assigning higher weights on the loss for the misclassified samples.
While these methods improve the face recognition performance in general datasets~\cite{lfw, cfpfp}, they still show considerably low performance in verifying child-adult pairs.

Several age-invariant face recognition models~\cite{mtl, DAL, AECNN, OECNN, look-across-elapse} try to tackle such an issue. 
AE-CNN~\cite{AECNN} and OE-CNN~\cite{OECNN} add an auxiliary age classifier to an identity classifier in order to eliminate the age-related information from the feature vectors. 
Wang~\etal present DAL~\cite{DAL} to disentangle the feature vectors into age and identity vectors through an adversarial learning approach.
While these approaches focus on obtaining the age-invariant identity features by modifying the architecture of classification models, Huang~\etal propose MTLFace~\cite{mtl} which leverages generative models and uses the representation obtained from the generative models for the face recognition. 
Although the methods mentioned above improve the performance in age-invariant face recognition, they bear several limitations.
First, they heavily rely on the additionally collected cross-age images.
This can not be a general solution since their performance is only guaranteed with the newly built train sets, which are currently not publicly available.
Second, introducing an additional generative model increases the training time and the computational cost~\cite{mtl}.
In this regard, we propose a \emph{simple yet effective method} which does not require additional child images or learnable parameters.

\vspace{-0.5cm}
\section{Analysis on the Inter-class Similarity}
\label{sec:motivation}
\vspace{-0.2cm}

\begin{figure}[t!]
  \centering
    \includegraphics[width=0.6\textwidth, clip]{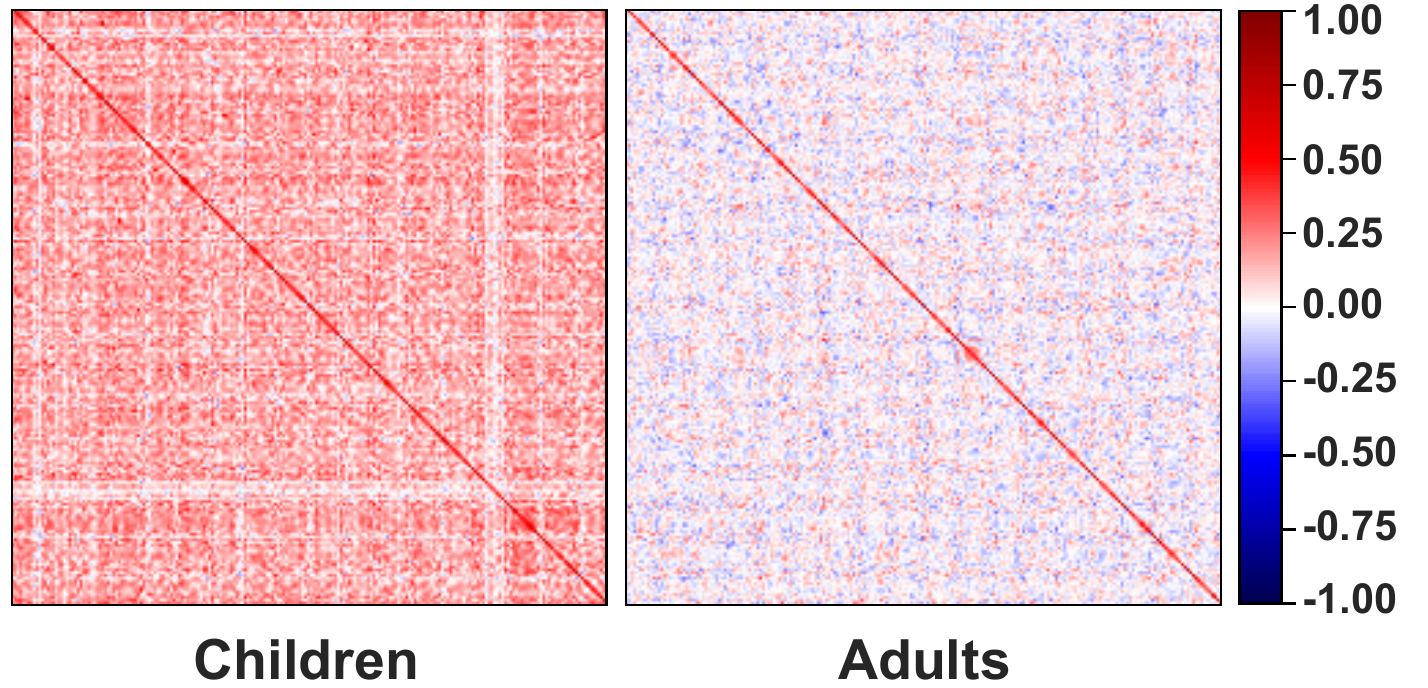}
    \vspace{-0.3cm}
    \caption{Intra- and inter-class similarity of child and adult images. The diagonal and off-diagonal cells indicate the intra- and inter-class similarity, respectively.}
    \label{fig:motivation}
    \vspace{-0.6cm}
\end{figure}

In order to examine the representation of child images, we analyze the intra-class similarity and the inter-class similarity of child and adult images in the CASIA-WebFace dataset~\cite{casia} using a pretrained ArcFace~\cite{arcface} model.
Since the CASIA-WebFace dataset does not include age labels for each image, we utilize the pseudo labels provided by Huang~\etal~\cite{mtl} who annotated the labels using the public Azure Facial API~\cite{azure}.
Inspired by Wang~\etal~\cite{DAL}, we group the pseudo age labels into 8 ranges: 0-12, 13-18, 19-25, 26-35, 36-45, 46-55, 56-65, and $\geq$ 66. 
We empirically observe the images and define the images in the age group of 0-12 as child images considering their facial appearances.
The distribution of the pseudo-labeled ages in CASIA-Webface dataset~\cite{casia} is presented in Fig.~\ref{fig:casia_agedist}.

Given the number of identities $n$ and the dimension of a feature vector $d$, the weight matrix of the last fully-connected layer in a classification model $\mathbf{W} \in \mathbb{R}^{d\times n}$ can be interpreted as a set of vectors that represent each class (\ie an identity)~\cite{sphereface, cosface}.
As aforementioned in Section.~\ref{sec:related-work}, we term such representative vector of the $i$-th identity in the last fully-connected layer $\mathbf{W}$ as \emph{prototype vector}, formulated as $\mathbf{W}_{i} \in \mathbb{R}^{d\times 1}$.
In this paper, the \emph{intra-class similarity} of an identity is defined as the averaged cosine similarity between feature vectors extracted from images and their corresponding prototype vectors, which is formulated as 
\vspace{-0.2cm}
\begin{equation}
\label{eq:intra}
    \mathbf{S}_{i}^{\text{intra}} = \frac{1}{N_{i}}\sum_{k=1}^{N_{i}}\mathbf{W}_i^\intercal f\left(\mathbf{x}_{i_{k}}\right),
\end{equation}
where $N_i$ is the number of training images in the $i$-th identity, $\mathbf{x}_{i_{k}}$ is the $k$-th input image of the $i$-th identity, and $f\left(\mathbf{x}_{i_{k}}\right)$ indicates the extracted feature vectors of $\mathbf{x}_{i_{k}}$. Also, \emph{Inter-class similarity} between two identities refers to the averaged cosine similarity between feature vectors extracted from images of different classes, which is described as 
\vspace{-0.2cm}
\begin{equation}
    \mathbf{S}_{ij}^{\text{inter}} = \frac{1}{N_i N_j} \sum_{k=1}^{N_i} \sum_{l=1}^{N_j} f(\mathbf{x}_{i_{k}})^\intercal f(\mathbf{x}_{j_{l}}), 
    \;\text{where}\;
    i\neq j.
\end{equation}
Fig.~\ref{fig:motivation} visualizes the averaged the intra- and inter-class similarity using child and adult images, respectively.
Each row and column indicate the index of 200 randomly sampled identities, respectively.
The diagonal cells indicate the intra-class similarity.
Each off-diagonal cell indicates the inter-class similarity of all image pairs between two given identities (\ie each from the row and the column, respectively).

There are two observations from the visualization. 
We observe that the intra-class similarity is both high in child and adult images (red-colored diagonals). 
This indicates that the ArcFace~\cite{arcface} alone can properly increase the intra-class similarity during the training phase.
On the other hand, the inter-class similarities between child images are substantially higher (colored as red) than those of adult images (colored as white or blue).
Such visualization indicates that the model did not fully learn to distinguish children.
This makes it challenging for the model to extract child representation, which results low performance in verifying child-adult pairs.
The feature vectors used for the visualization are obtained from the training dataset, so the inter-class similarities should have been minimized since they are explicitly supervised from the ArcFace loss. 
However, we observe that the inter-class similarities between child images are still high.
Based on the two observations, despite the high intra-class similarity achieved by the model, we conclude that the poor performance on child-adult pairs is derived from the lack of discriminative features obtained from child images due to the high inter-class similarity.
In this regard, we propose a novel loss function called Inter-Prototype loss to mitigate this issue. 

\vspace{-0.4cm}
\section{Proposed Approach}
\label{sec:method}

\subsection{Margin-based loss}
Margin-based loss aims to increase the cosine similarity between feature vectors and their corresponding prototype vectors by utilizing margin terms, which penalizes the angles between them.
The margin-based loss function we use throughout our experiment is the ArcFace~\cite{arcface}, which is written as follows,
\vspace{-0.2cm}
\begin{equation}
    \mathcal{L}_m = \frac{1}{N}\sum_{i=1}^{N} - \log\frac{e^{s\;(\cos(\theta_{y_i, i} \; +m))}}{e^{s\;(\cos(\theta_{y_i, i} \; +m))} + \sum_{j=1,j\neq y_i}^{n}e^{s\;\cos{\theta_{j,i}}}},
\end{equation}
\vspace{-0.2cm}
\begin{equation}
    \cos{\theta_{ji}} = \frac{\mathbf{W}_j^\intercal f(\mathbf{x}_i)} {\mathopen\| \mathbf{W}_j\mathclose\| \mathopen\| f(\mathbf{x}_i)\mathclose\|},
\end{equation}
where $N$ is the batch size, $n$ is the number of classes, $y_i$ is the ground-truth class, $\cos\theta_{j,i}$ is the cosine similarity between $\mathbf{W}_j$ and $x_i$, $s$ is the scale factor, and $m$ is the margin.

\vspace{-0.4cm}
\subsection{Inter-Prototype loss}
\begin{figure}[t!]
  \centering
    \includegraphics[width=\textwidth, clip]{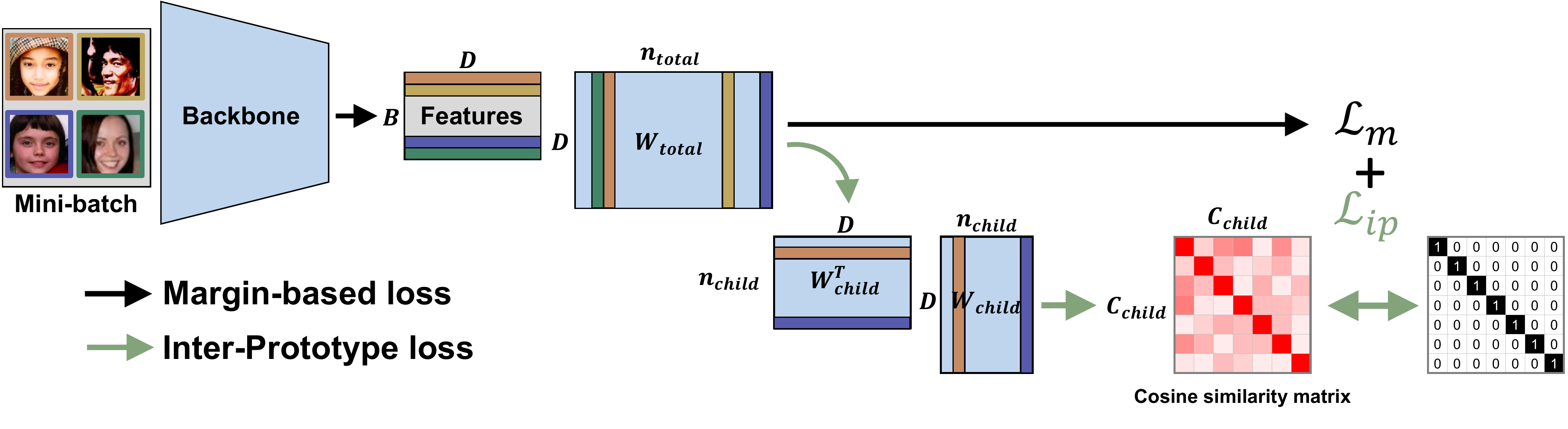}
    \vspace{-0.9cm}
    \caption{Overview of our method. We add our Inter-Prototype loss to an existing margin-based loss function~\cite{arcface}. Among all identities, we obtain the identities which include child images ($n_{child}$). Then, we compute a cosine similarity matrix of the prototype vectors using those identities. We enforce the cosine similarity matrix of the prototype vectors to become an identity matrix.}
    \vspace{-0.4cm}
    \label{fig:overview}
\end{figure}

The Inter-Prototype loss is not imposed on the feature vectors extracted from the backbone model, but rather enforced on the prototype vectors.
Fig.~\ref{fig:overview} illustrates the workflow of our method.
We collect the prototype vectors of classes that contain the child images denoted as $\mathbf{W}_{child} \in \mathbb{R}^{d\times n_{child}}$, where $n_{child}$ is the number of classes containing the child images.

After normalizing the prototype vectors of $\mathbf{W}_{child}$, we obtain the cosine similarity matrix $\mathbf{C} \in \mathbb{R}^{n_{child} \times n_{child}}$ using the prototype vectors.
The Inter-Prototype loss focuses on minimizing the inter-class similarity between identities including child images which is formulated as,
\vspace{-0.15cm}
\begin{equation}
\label{eq:iploss}
\mathcal{L}_{ip} = \sum_{i=1}^{n_{child}} \sum_{j\neq i} \mathbf{C}_{ij}^2,
\end{equation}
\begin{equation}
\label{eq:cmatrix}
\mathbf{C}_{ij} = \frac{\mathbf{W}_{child_i}^\intercal \mathbf{W}_{child_j}}{\mathopen\| \mathbf{W}_{child_i} \mathclose\| \mathopen\| \mathbf{W}_{child_j}\mathclose\|}, \mathbf{W}_{child_i} \in \mathbb{R}^{d\times1}.
\end{equation}
Since we add the Inter-Prototype loss on the margin-based loss function~\cite{arcface},
our final objective function is described as
\begin{equation}
\label{eq:total}
\mathcal{L}_{total} = \mathcal{L}_{m} + \lambda_{ip} \mathcal{L}_{ip},
\end{equation}
where $\mathcal{L}_{m}$ refers to a margin-based loss function~\cite{arcface, cosface, sphereface} and $\lambda_{ip}$ is a hyper-parameter to balance between the two loss functions which is set to 1. 
One possible solution for reducing the inter-class similarity between the child identities would be directly sampling the child image pairs and reducing their similarities in the embedding space.
However, this approach requires sampling additional child image pairs in a single batch, which is sensitive to the sampling.
Even without such additional sampling of child images, adding the Inter-Prototype effectively minimizes the inter-class similarities between child images.

Algorithm~\ref{alg:main} explains the overall training procedure of our approach with the pseudo code.
\begin{figure}[b!]
\centering
\scalebox{0.9}{
\begin{minipage}{0.8\linewidth}
\begin{algorithm}[H]
\label{alg:main}
\caption{Training with Inter-Prototype Loss}
\SetAlgoLined
\KwIn{image $x$, label $y$, iteration $i$, learning rate $\alpha$, margin-based loss $\mathcal{L}_m$, Inter-Prototype loss $\mathcal{L}_{ip}$, identity matrix $\mathbf{I}$, backbone network $f(\cdot)$, backbone network parameters $\Theta$, and prototype vectors $\mathbf{W}$} 
 \While{\textit{not converged}}{
  \begin{varwidth}{\linewidth}
        Extract feature vectors $f(x_i)$ \par
        Calculate the cosine similarity matrix 
        $\mathbf{C}_{} = \frac{\mathbf{W}_{child}^\intercal \mathbf{W}_{child}}{\mathopen\| \mathbf{W}_{child} \mathclose\| \mathopen\| \mathbf{W}_{child}\mathclose\|}$ \par
        Compute the loss $\mathcal{L}_{total} \text{= } \mathcal{L}_{m}(f(x_i), y) + \lambda_{ip}\mathcal{L}_{ip}(\mathbf{C}, \mathbf{I})$ (Eq.~\ref{eq:total}) \par
        Update $\Theta$ and $\mathbf{W}$ by: \par

        \vspace{0.1cm}
        \enspace $\mathbf{W}^{i+1} \text{= } \mathbf{W}^{i} \text{ - } \alpha\frac{\partial\mathcal{L}_{total}}{\partial\mathbf{W}^{i}}$ where $ \frac{\partial\mathcal{L}_{total}}{\partial\mathbf{W}^{i}}$ = $\frac{\partial\mathcal{L}_{m}}{\partial\mathbf{W}^{i}} + \lambda_{ip} \frac{\partial\mathcal{L}_{ip}}{\partial\mathbf{W}^{i}},$ \par
        \vspace{0.1cm}
        \enspace $\Theta^{i+1} \text{= } \Theta^{i} \text{ - } \alpha \frac{\partial\mathcal{L}_{total}}{\partial{ \Theta^{i}}}$ \par
  \end{varwidth}}
\end{algorithm}
\end{minipage}}
\end{figure}
The Inter-Prototype loss improves the face recognition accuracy on child-adult pairs in two aspects.
First, we can improve the capability of our backbone network to distinguish the difference between children.
The high inter-class similarity between different child images indicates that the model is not able to discriminate between two different children, implying that the representation of child images is not formed to a reasonable level.
The Inter-Prototype loss addresses such an issue by explicitly minimizing the inter-class similarity between identities including the child images, improving the representation of child images overall. 
Second, adding the Inter-Prototype loss also minimizes the inter-class similarity between \emph{child and adult} images of different identities. 
The prototype vector used for the Inter-Prototype loss is a representative vector for both child and adult images of a given identity. 
Since we supervise such prototype vectors to be distinct from each other, we are also minimizing the inter-class similarity between child and adult pairs of different identities, as shown in Fig.~\ref{fig:discussion-visualization}.
Additional to such intuition, Inter-Prototype loss is applicable to existing margin-based models as long as the last fully-connected layer can be interpreted as a set of prototype vectors.
Moreover, it is insensitive to the number of child images in a mini-batch and does not require additional learnable parameters.

\vspace{-0.5cm}
\section{Experiments}
\vspace{-0.2cm}
\subsection{Datasets}
For the training dataset, we separately conduct experiments using the CASIA-WebFace dataset~\cite{casia} and the MS1MV2 dataset in order to evaluate our proposed method both in the small and the large dataset. 
The CASIA-WebFace dataset consists of 0.5M images for 10K individuals (\ie small evaluation protocol) while the MS1MV2 dataset includes 5M images for 85K individuals (\ie large evaluation protocol).
As aforementioned, we utilize the age labels annotated from MTLFace~\cite{mtl} for both datasets to specify child images.
We could not train our model with the newly built datasets with additional child images, as done in the previous age-invariant face recognition studies~\cite{OECNN, DAL, mtl}, since they are not publicly available.

Face verification is the task of determining whether a pair of facial images belong to the same identity or not. 
For the face verification with child-adult pairs, we build the child-adult pairs using the original cross-age datasets, FG-NET~\cite{fgnet} and AgeDB~\cite{agedb}.
We pair the images of children (\ie under age 13) and adults with an age gap of more than 20 years and 30 years.
We name these datasets as FG-NET-C20, AgeDB-C20, FG-NET-C30, and AgeDB-C30 in our work. 
The LAG dataset~\cite{lag}, a publicly available test dataset which is composed of images with large age gaps, is also used for the evaluation.
Since the LAG dataset does not have age labels, we do not consider the age gaps for pairing the images.
All test sets have the equal number of positive and negative pairs.
We also compare our approach with previous age-invariant face recognition models~\cite{OECNN, DAL, mtl} using the existing general and cross-age face recognition test sets.
Note that the difference between our test datasets and the existing cross-age test datasets is that we \textbf{intentionally include a child image in each pair} while the latter does not necessarily include them. 

Face identification is the task of predicting the identity of a given facial image (\ie probe image) among the candidate images (\ie gallery images). 
To be more specific, we use the child image as the probe image and use the adult images as gallery images.
We evaluate models with the rank-1 accuracy. 
We select child-adult pairs for each identity from the AgeDB-20C, AgeDB-30C dataset, and LAG dataset to build an identification set.
We did not use FG-NET since the number of child-adult pairs is insufficient for the evaluation.
Further details of the test datasets are included in our supplementary material.

\begin{table}[t!]
\centering
\scalebox{0.8}{
\begin{tabular}{cccccc} \hline
Method               & AgeDB-C20 & FG-NET-C20 & AgeDB-C30 & FG-NET-C30 & LAG~\cite{lag} \\ \hline
CurricularFace~\cite{curricular}   & 82.11\stdv{0.36} & 81.56\stdv{0.56} & 81.49\stdv{0.42} & 78.85\stdv{0.16} & 90.35\stdv{0.06}  \\
MV-AM~\cite{MV-softmax}            & 81.01\stdv{0.17} & 81.37\stdv{0.90} & 80.45\stdv{0.78} & 79.48\stdv{1.02} & 89.38\stdv{0.15} \\
BroadFace~\cite{broadface}         & 82.93\stdv{0.20} & 80.4\stdv{0.25}  & 81.85\stdv{0.48} & 79.03\stdv{0.54} & 89.63\stdv{0.26} \\
CosFace~\cite{cosface}             & 81.65\stdv{0.66} & 81.27\stdv{0.57} & 81.02\stdv{0.79} & 79.03\stdv{1.42} & 89.58\stdv{0.06} \\
ArcFace~\cite{arcface}             & 82.65\stdv{0.31} & 80.37\stdv{1.05} & 81.91\stdv{0.55} & 79.03\stdv{0.94} & 89.80\stdv{0.25} \\ 
OECNN~\cite{OECNN}                 & 78.87\stdv{0.62} & 78.95\stdv{0.70} & 78.24\stdv{0.86} & 77.78\stdv{0.82} & 87.33\stdv{0.49} \\
DAL~\cite{DAL}                     & 77.28\stdv{1.41} & 77.18\stdv{1.19} & 75.90\stdv{1.01} & 76.53\stdv{2.50} & 86.13\stdv{0.51} \\
MTL~\cite{mtl}                     & 82.18\stdv{0.47} & 80.42\stdv{0.29} & 80.64\stdv{1.17} & 80.02\stdv{0.68} & 88.17\stdv{0.32} \\ \hline \hline
Ours        & \textbf{83.55}\stdv{0.20} & \textbf{82.27}\stdv{1.03} & \textbf{82.97}\stdv{0.76} & \textbf{81.00}\stdv{1.27} & \textbf{90.49}\stdv{0.17} \\ \hline
\end{tabular}}
\vspace{0.25cm}
\caption{Face verification accuracy (\%) trained on CASIA-WebFace~\cite{casia} and evaluated on test sets with child-adult pairs.}
\vspace{-0.2cm}
\label{tab:verification_child_s}
\end{table}

\begin{table}[t!]
\centering
\scalebox{0.8}{
\begin{tabular}{cccccc} \hline
Method                             & AgeDB-C20        & FG-NET-C20       & AgeDB-C30        & FG-NET-C30       & LAG~\cite{lag} \\ \hline
CurricularFace~\cite{curricular}   & 85.76\stdv{1.34} & 83.03\stdv{1.70} & 84.85\stdv{1.60} & 80.46\stdv{1.66} & 91.47\stdv{0.86}  \\
MV-AM~\cite{MV-softmax}            & 84.70\stdv{0.91} & 81.72\stdv{1.27} & 83.43\stdv{0.51} & 79.12\stdv{1.09} & 90.81\stdv{0.68} \\
CosFace~\cite{cosface}             & 85.82\stdv{0.03} & 81.46\stdv{0.62} & 85.07\stdv{0.50} & 78.05\stdv{0.86} & 91.37\stdv{0.23} \\
ArcFace~\cite{arcface}             & 86.21\stdv{0.67} & 83.88\stdv{0.91} & 85.85\stdv{0.60} & 80.91\stdv{1.17} & 91.78\stdv{0.34} \\ 
OECNN~\cite{OECNN}                 & 83.94\stdv{0.42} & 84.88\stdv{1.12} & 82.81\stdv{0.69} & 83.96\stdv{1.12} & 90.33\stdv{1.03} \\
DAL~\cite{DAL}                     & 83.58\stdv{0.10} & 82.11\stdv{1.17} & 82.59\stdv{0.36} & 82.62\stdv{1.35} & 90.31\stdv{0.46} \\ \hline \hline
Ours                               & \textbf{87.36}\stdv{0.08} & \textbf{85.81}\stdv{0.24} & \textbf{87.01}\stdv{0.15} & \textbf{85.04}\stdv{0.56} & \textbf{93.12}\stdv{0.23} \\ \hline
\end{tabular}}
\vspace{0.25cm}
\caption{Face verification accuracy (\%) trained on MS1MV2 and evaluated on test sets with child-adult pairs.}
\vspace{-0.3cm}
\label{tab:verification_child_l}
\end{table}

\vspace{0.1cm}
\begin{table}[t!]
\centering
\scalebox{0.8}{
\begin{tabular}{cccc|cc} \hline
Method                           & AgeDB-C20 & AgeDB-C30 & LAG~\cite{lag} & FLOPs & \# of params. \\ \hline
OECNN~\cite{OECNN}               & 31.75\stdv{4.77} & 24.19\stdv{3.23} & 42.62\stdv{1.69} & 6.32B & 49.21M \\
DAL~\cite{DAL}                   & 27.51\stdv{4.57} & 22.58\stdv{5.81} & 38.77\stdv{1.53} & 6.32B & 50.26M \\
MTL~\cite{mtl}                   & 33.86\stdv{2.42} & 31.73\stdv{4.93} & 48.76\stdv{0.75} & 12.51B & 134.05M \\ \hline \hline
Ours                             & \textbf{41.80}\stdv{0.92} & \textbf{34.94}\stdv{0.93} & \textbf{53.70}\stdv{0.42}  & \bf{6.32B} & \bf{43.79M} \\ \hline
\end{tabular}}
\vspace{0.2cm}
\caption{Face identification accuracy (Rank-1) (\%) on test sets with along with computational costs (FLOPs, number of parameters) between age-invariant face recognition models. B and M refer to billion and million, respectively.}
\vspace{-0.2cm}
\label{tab:identification_child}
\end{table}

\vspace{-0.4cm}
\subsection{Implementation Details}
For the data processing, we leverage RetinaFace~\cite{retinaface} to detect faces and facial landmarks on both train and test datasets. After the face detection and alignment, we perform similarity transformation according to the five facial key points (two eyes, nose, and two mouth corners) and obtain images of 112$\times$112 pixels.
We utilize ResNet50~\cite{resnet} as the backbone network and obtain 512-dimensional embedded feature vectors for the evaluation.
We use ArcFace\cite{arcface}, with a margin of 0.5 and a scale factor of 64, for $\mathcal{L}_m$ in our method.
We train our model for 50 epochs using a batch size of 64.
We use the SGD optimizer~\cite{sgd} with a weight decay of 0.0005 and the learning rate is initially set to 0.1, divided by 10 at the epoch of 28, 36, 46. 
For each experiment, we report the averaged accuracy and its standard deviation over three independent trials.
For the Inter-Prototype loss, we obtain ($n_{child}$, $n_{total}$)=($229$, $10,575$) and ($1,148$, $85,742$) in the CASIA-WebFace dataset~\cite{casia} and MS1MV2, respectively.

\vspace{-0.2cm}
\subsection{Quantitative Evaluation}
\noindent \textbf{Face verification/identification with child-adult pairs}
Table.~\ref{tab:verification_child_s} and Table.~\ref{tab:verification_child_l} show that our approach outperforms all baseline approaches on every test set in the face verification with child-adult pairs when trained with the small and large dataset, respectively. 
Our proposed approach shows its strength as the age gap between verification pairs increases from at least 20 to 30.
Additionally, recent age-invariant face recognition models, OECNN~\cite{OECNN}, DAL~\cite{DAL}, and MTL~\cite{mtl}, generally show low verification accuracy across all test sets.
We believe this result demonstrates that previous studies heavily rely on the additionally collected child images in the training dataset.
Table.~\ref{tab:identification_child} also shows that our approach outperforms the existing work of age-invariant face recognition on the face identification using three test sets with child-adult pairs.
Along with the outstanding face recognition performances, we also report the FLOPs (\ie the number of floating-point operations) and the required number of parameters for each method.
Note that MTL~\cite{mtl} requires a considerable amount of computational cost since it utilizes an image generation model. 
This shows that using image generation models requires lengthy training and a substantial amount of parameters.
This result again demonstrates that our method is \emph{simple yet effective}.

\begin{table}[t!]
\centering
\scalebox{0.8}{
\begin{tabular}{cccccc} \hline
Method                    & LFW & CFP-FP & CPLFW & AgeDB-30 & CALFW \\ \hline
ArcFace~\cite{arcface}    & \textbf{99.53}$^\text{*}$ & \textbf{95.56}$^\text{*}$  & 85.79\stdv{0.48} & \textbf{95.15}$^\text{*}$ & 92.89\stdv{0.29} \\
OECNN~\cite{OECNN}        & 98.65\stdv{0.20}  & 91.81\stdv{0.85} & 84.31\stdv{0.56} & 91.82\stdv{0.38} & 91.28\stdv{0.27} \\
DAL~\cite{DAL}            & 98.37\stdv{0.29}  & 90.55\stdv{1.12} & 84.27\stdv{0.22} & 90.24\stdv{1.06} & 90.39\stdv{0.48} \\
MTL~\cite{mtl}            & 99.04\stdv{0.17} & 92.67\stdv{0.79} & 85.23\stdv{0.55} & 92.95\stdv{0.26} & 92.34\stdv{0.16} \\ \hline \hline
Ours                      & 99.24\stdv{0.15} & 94.71\stdv{0.37}  & \textbf{86.62}\stdv{0.86} & 94.19\stdv{0.31}  & \textbf{93.33}\stdv{0.15} \\ \hline
\end{tabular}}
\vspace{0.25cm}
\caption{Face verification accuracy (\%) with general/cross-age datasets. $^\text{*}$ indicates that the results are from the original paper of ArcFace~\cite{arcface}. Note that the original paper only reports the best accuracy while we report the averaged accuracy of three independent trials.}
\vspace{-0.4cm}
\label{tab:verification_general}
\end{table}

\vspace{0.2cm}
\noindent \textbf{Face verification with general/cross-age test sets}
Table.~\ref{tab:verification_general} shows the face recognition performance on both general and cross-age test sets. 
Since both our approach and the previous age-invariant face recognition models focus on the pairs with age gaps, there may exist performance degradation on the general test sets.  
To be specific, compared to Arcface~\cite{arcface}, the previous age-invariant face recognition models show large performance degradation in general and cross-age test sets. 
On the other hand, our approach shows reasonable performance in general and cross-age test sets even with superior performances in the child-adult pairs.
This result demonstrates that the Inter-Prototype loss does not bring negative effect on the general face recognition performance compared to existing age-invariant face recognition models.

\vspace{-0.4cm}
\subsection{Qualitative Analysis}
\begin{figure}[t!]
    \centering
    \vspace{-0.3cm}
    \includegraphics[width=\textwidth, clip]{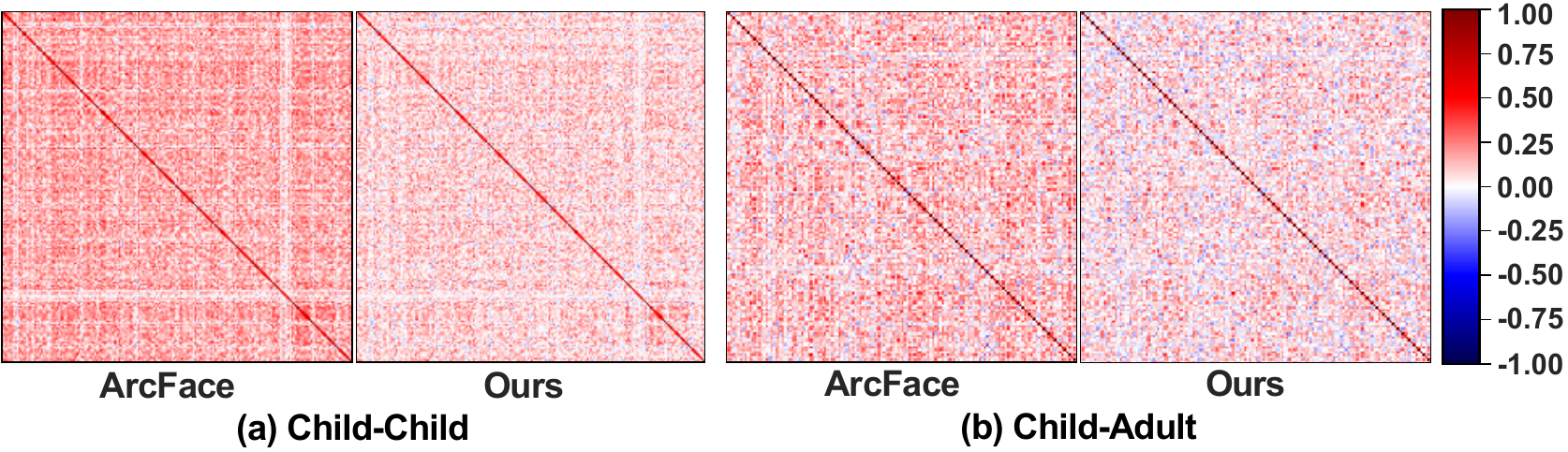}
    \vspace{-0.6cm}
    \caption{Comparison of inter-class similarity between ArcFace and our approach using (a) child-child pairs and (b) child-adult pairs. Each row and column indicate the identities used for the visualization, respectively.}
    \vspace{-0.2cm}
    \label{fig:discussion-visualization}
\end{figure}

\noindent \textbf{Reduced inter-class similarity}
Fig.~\ref{fig:discussion-visualization} (a) and Fig.~\ref{fig:discussion-visualization} (b) visualize the inter-class similarity using the child-child pairs and child-adult pairs on ArcFace~\cite{arcface} and our approach, respectively. 
We use the identities that include child images for the visualization.
The inter-class similarity between child-child pairs is generally high in ArcFace, overall colored red.
After adding the Inter-Prototype loss, the inter-class similarity decreases substantially, showing reduced portions of red, which demonstrates the effectiveness of our proposed approach. 
The Inter-Prototype loss also decreases the inter-class similarity between child-adult pairs. 
As aforementioned in Section.~\ref{sec:motivation}, the Inter-Prototype loss also enforces the child-adult pairs of different identities to be distinct. 
Due to this fact, we observe that the off-diagonals of our approach in Fig.~\ref{fig:discussion-visualization} (b) are colored blue or white compared to the ArcFace.

\vspace{0.4cm}
\begin{table}[t!]
\centering
\scalebox{0.85}{
\begin{tabular}{cccccc} \hline
Method                      & AgeDB-C20 & FG-NET-C20 & AgeDB-C30 & FG-NET-C30 & LAG~\cite{lag} \\ \hline
ArcFace~\cite{arcface}      & 82.65\stdv{0.31} & 80.37\stdv{1.05} & 81.91\stdv{0.55} & 79.03\stdv{0.94} & 89.80\stdv{0.25} \\ 
Ours + Full                 & 83.29\stdv{0.39} & 82.11\stdv{0.44} & 82.41\stdv{0.40} & 80.56\stdv{0.67} & 90.44\stdv{0.08} \\  
Ours  + Child               & \textbf{83.55}\stdv{0.20} & \textbf{82.27}\stdv{1.03} & \textbf{82.97}\stdv{0.76} & \textbf{81.00}\stdv{1.27} & \textbf{90.49}\stdv{0.17} \\ \hline
\end{tabular}}
\vspace{0.25cm}
\caption{Face verification accuracy (\%) on the different number of prototype vectors. Ours~+~Full indicates applying the Inter-Prototype loss on all identities. Similarly, Ours~+~Child refers to applying the Inter-Prototype loss only on identities including the child images.}
\vspace{-0.6cm}
\label{tab:ablation_full}
\end{table}

\vspace{-0.6cm}
\begin{table}[t!]
\centering
\scalebox{0.85}{
\begin{tabular}{cccc} \hline
Method                            & AgeDB-C20 & AgeDB-C30 & LAG~\cite{lag} \\ \hline
CosFace~\cite{cosface}             & 81.65\stdv{0.66} & 81.02\stdv{0.79} & 89.58\stdv{0.06} \\
Ours + CosFace                     & \textbf{82.41}\stdv{0.57} & \textbf{81.52}\stdv{0.57} & \textbf{90.14}\stdv{0.56} \\ \hline \hline
ArcFace~\cite{arcface}             & 82.65\stdv{0.31} & 81.91\stdv{0.55} & 89.80\stdv{0.25} \\ 
Ours + ArcFace                     & \textbf{83.55}\stdv{0.20} & \textbf{82.97}\stdv{0.76} & \textbf{90.49}\stdv{0.17} \\ \hline
\end{tabular}}
\vspace{0.1cm}
\caption{Face verification accuracy (\%) of adding Inter-Prototype loss on both CosFace and Arcface.}
\vspace{0.1cm}
\label{tab:verification_ablation}
\end{table}

\begin{figure}[t!]
    \centering
    \includegraphics[width=\textwidth, clip]{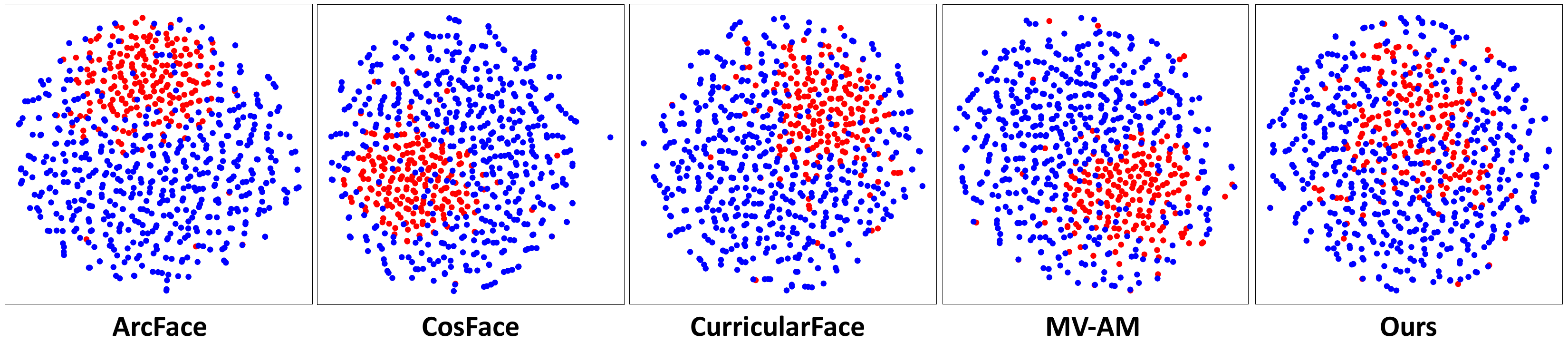}
    \caption{2D projection of prototype vectors. Red and blue indicate the prototype vectors that include the child images and the ones without them, respectively.}
    \vspace{-0.5cm}
    \label{fig:discussion-visualization2d}
\end{figure}

\vspace{0.4cm}
\noindent \textbf{2D projection of representation}
Fig.~\ref{fig:discussion-visualization2d} visualizes the 2D projection of prototype vectors after training with ArcFace and our approach.
The red and blue dots indicate the embedded prototype vectors that include child images and the ones that do not, respectively.
The 2D projection of prototype vectors trained with existing face recognition models show that the identities including child images are embedded closely, indicating that the model does not completely distinguish the child images. 
We observe that adding the Inter-Prototype loss mitigates this issue by inspecting that the red dots are evenly distributed in the embedding space. 

\vspace{-0.4cm}
\section{Discussion}
\noindent \textbf{Applicability on margin-based loss functions}
Table.~\ref{tab:verification_ablation} demonstrates that the proposed Inter-Prototype loss improves the face verification accuracy of both CosFace and ArcFace on AgeDB-C20, AgeDB-C30, and LAG.
As described in Eq.~\ref{eq:total}, we add the Inter-Prototype loss to the margin-based loss functions (\eg CosFace and ArcFace), respectively, and show that adding it improves the face verification accuracy in each model.
Such a result demonstrates that the proposed Inter-Prototype loss can be widely applied to existing margin-based loss functions as long as the last fully-connected layer can be interpreted as a set of prototype vectors.

\vspace{0.3cm}
\noindent \textbf{Extending the Inter-Prototype loss on all identities}
Table~\ref{tab:ablation_full} compares the face verification performance of applying the Inter-Prototype loss on all identities and only the identities including child images. 
Applying the Inter-Prototype loss on all identities improves the performance of ArcFace. 
This indicates that even without using the age labels, minimizing the inter-class similarity between identities improves the representation of child images.
However, we observe that using only the identities including child images shows further improvement.
We conjecture that applying the Inter-Prototype loss on all identities gives less emphasis on the loss backpropagated on the prototype vectors including child images. 
Therefore, this experiment shows that the Inter-Prototype loss effectively minimizes the inter-class similarity when given a specific set of identities.

\vspace{-0.3cm}
\section{Conclusion}
\vspace{-0.3cm}
In this work, we first reveal that reducing the inter-class similarity among child images is crucial for verifying child-adult pairs.
Based on our observation, we propose a novel loss function called the Inter-Prototype loss to minimize the inter-class similarity between child images using the prototype vectors.
Our loss function does not require additional child images or learnable parameters which were necessary for previous age-invariant face recognition models. 
Through extensive experiments, we demonstrate that our method clearly outperforms the existing face recognition models on test sets with child-adult pairs. 
Additionally, the various visualization and ablation studies verify the effectiveness of our approach.
We hope our simple yet effective method inspires the future researchers who plan to delve into this area.

\vspace{-0.2cm}
\section{Acknowledgement}
This work was supported by the Institute of Information \& communications Technology Planning \& Evaluation (IITP) grant funded by the Korean government(MSIT) (No. 2019-0-00075, Artificial Intelligence Graduate School Program(KAIST)), the National Research Foundation of Korea (NRF) grant funded by the Korean government (MSIT) (No. NRF-2018M3E3A1057305, No. NRF-2019R1A2C4070420), and Kakao Enterprise.

\clearpage
\bibliography{egbib}

\begin{thebibliography}{26}
\providecommand{\natexlab}[1]{#1}
\providecommand{\url}[1]{\texttt{#1}}
\expandafter\ifx\csname urlstyle\endcsname\relax
  \providecommand{\doi}[1]{doi: #1}\else
  \providecommand{\doi}{doi: \begingroup \urlstyle{rm}\Url}\fi

\bibitem[Albiero and Bowyer(2020)]{sexist}
V{\'{\i}}tor Albiero and Kevin~W. Bowyer.
\newblock Is face recognition sexist? no, gendered hairstyles and biology are.
\newblock In \emph{31st British Machine Vision Conference 2020, {BMVC} 2020,
  Virtual Event, UK, September 7-10, 2020}. {BMVA} Press, 2020.

\bibitem[Azure(2021)]{azure}
Microsoft Azure.
\newblock Microsoft azure cognitive services facial recognition., 2021.
\newblock URL
  \url{https://azure.microsoft.com/en-us/services/cognitive-services/face/}.

\bibitem[Bianco(2017)]{lag}
Simone Bianco.
\newblock Large age-gap face verification by feature injection in deep
  networks.
\newblock \emph{Pattern Recognit. Lett.}, 90:\penalty0 36--42, 2017.

\bibitem[Deng et~al.(2019)Deng, Guo, Xue, and Zafeiriou]{arcface}
Jiankang Deng, Jia Guo, Niannan Xue, and Stefanos Zafeiriou.
\newblock Arcface: Additive angular margin loss for deep face recognition.
\newblock In \emph{{IEEE} Conference on Computer Vision and Pattern
  Recognition, {CVPR} 2019, Long Beach, CA, USA, June 16-20, 2019}, pages
  4690--4699. Computer Vision Foundation / {IEEE}, 2019.

\bibitem[Deng et~al.(2020)Deng, Guo, Ververas, Kotsia, and
  Zafeiriou]{retinaface}
Jiankang Deng, Jia Guo, Evangelos Ververas, Irene Kotsia, and Stefanos
  Zafeiriou.
\newblock Retinaface: Single-shot multi-level face localisation in the wild.
\newblock In \emph{2020 {IEEE/CVF} Conference on Computer Vision and Pattern
  Recognition, {CVPR} 2020, Seattle, WA, USA, June 13-19, 2020}, pages
  5202--5211. {IEEE}, 2020.

\bibitem[Fu et~al.(2014)Fu, Hospedales, Xiang, Yao, and Gong]{fgnet}
Yanwei Fu, Timothy~M. Hospedales, Tao Xiang, Yuan Yao, and Shaogang Gong.
\newblock Interestingness prediction by robust learning to rank.
\newblock In \emph{ECCV}, 2014.

\bibitem[Guo et~al.(2016)Guo, Zhang, Hu, He, and Gao]{ms1mv2}
Yandong Guo, Lei Zhang, Yuxiao Hu, Xiaodong He, and Jianfeng Gao.
\newblock Ms-celeb-1m: A dataset and benchmark for large-scale face
  recognition, 2016.

\bibitem[He et~al.(2016)He, Zhang, Ren, and Sun]{resnet}
Kaiming He, Xiangyu Zhang, Shaoqing Ren, and Jian Sun.
\newblock Deep residual learning for image recognition.
\newblock In \emph{2016 {IEEE} Conference on Computer Vision and Pattern
  Recognitionn, {CVPR} 2016, Las Vegas, NV, USA, June 27-30, 2016}, pages
  770--778. {IEEE} Computer Society, 2016.
\newblock \doi{10.1109/CVPR.2016.90}.

\bibitem[Hu et~al.(2018)Hu, Shen, and Sun]{squeeze}
Jie Hu, Li~Shen, and Gang Sun.
\newblock Squeeze-and-excitation networks.
\newblock In \emph{Proceedings of the IEEE Conference on Computer Vision and
  Pattern Recognition (CVPR)}, June 2018.

\bibitem[Huang et~al.(2007)Huang, Ramesh, Berg, and Learned-Miller]{lfw}
Gary~B. Huang, Manu Ramesh, Tamara Berg, and Erik Learned-Miller.
\newblock Labeled faces in the wild: A database for studying face recognition
  in unconstrained environments.
\newblock Technical Report 07-49, University of Massachusetts, Amherst, October
  2007.

\bibitem[Huang et~al.(2020)Huang, Wang, Tai, Liu, Shen, Li, Li, and
  Huang]{curricular}
Yuge Huang, Yuhan Wang, Ying Tai, Xiaoming Liu, Pengcheng Shen, Shaoxin Li,
  Jilin Li, and Feiyue Huang.
\newblock Curricularface: Adaptive curriculum learning loss for deep face
  recognition.
\newblock In \emph{2020 {IEEE/CVF} Conference on Computer Vision and Pattern
  Recognition, {CVPR} 2020, Seattle, WA, USA, June 13-19, 2020}. {IEEE}, 2020.

\bibitem[Huang et~al.(2021)Huang, Zhang, and Shan]{mtl}
Zhizhong Huang, Junping Zhang, and Hongming Shan.
\newblock When age-invariant face recognition meets face age synthesis: A
  multi-task learning framework.
\newblock In \emph{CVPR}, 2021.

\bibitem[Kim et~al.(2020)Kim, Park, and Shin]{broadface}
Yonghyun Kim, Wonpyo Park, and Jongju Shin.
\newblock Broadface: Looking at tens of thousands of people at once for face
  recognition.
\newblock In \emph{European Conference on Computer Vision}, pages 536--552,
  2020.

\bibitem[Liu et~al.(2017)Liu, Wen, Yu, Li, Raj, and Song]{sphereface}
Weiyang Liu, Yandong Wen, Zhiding Yu, Ming Li, Bhiksha Raj, and Le~Song.
\newblock Sphereface: Deep hypersphere embedding for face recognition.
\newblock In \emph{2017 {IEEE} Conference on Computer Vision and Pattern
  Recognition, {CVPR} 2017, Honolulu, HI, USA, July 21-26, 2017}, pages
  6738--6746. {IEEE} Computer Society, 2017.

\bibitem[Moschoglou et~al.(2017)Moschoglou, Papaioannou, Sagonas, Deng, Kotsia,
  and Zafeiriou]{agedb}
Stylianos Moschoglou, Athanasios Papaioannou, Christos Sagonas, Jiankang Deng,
  Irene Kotsia, and Stefanos Zafeiriou.
\newblock Agedb: the first manually collected, in-the-wild age database.
\newblock In \emph{Proceedings of the IEEE Conference on Computer Vision and
  Pattern Recognition Workshop}, page~5, 2017.

\bibitem[Ruder(2016)]{sgd}
Sebastian Ruder.
\newblock An overview of gradient descent optimization algorithms.
\newblock \emph{arXiv preprint arXiv:1609.04747}, 2016.

\bibitem[Sengupta et~al.(2016)Sengupta, Cheng, Castillo, Patel, Chellappa, and
  Jacobs]{cfpfp}
S.~Sengupta, J.C. Cheng, C.D. Castillo, V.M. Patel, R.~Chellappa, and D.W.
  Jacobs.
\newblock Frontal to profile face verification in the wild.
\newblock In \emph{IEEE Conference on Applications of Computer Vision}, 2016.

\bibitem[Taigman et~al.(2014{\natexlab{a}})Taigman, Yang, Ranzato, and
  Wolf]{deepface}
Yaniv Taigman, Ming Yang, Marc'Aurelio Ranzato, and Lior Wolf.
\newblock Deepface: Closing the gap to human-level performance in face
  verification.
\newblock In \emph{2014 IEEE Conference on Computer Vision and Pattern
  Recognition}, pages 1701--1708, 2014{\natexlab{a}}.

\bibitem[Taigman et~al.(2014{\natexlab{b}})Taigman, Yang, Ranzato, and
  Wolf]{deepface_cvpr}
Yaniv Taigman, Ming Yang, Marc'Aurelio Ranzato, and Lior Wolf.
\newblock Deepface: Closing the gap to human-level performance in face
  verification.
\newblock In \emph{2014 IEEE Conference on Computer Vision and Pattern
  Recognition}, pages 1701--1708, 2014{\natexlab{b}}.
\newblock \doi{10.1109/CVPR.2014.220}.

\bibitem[Wang et~al.(2018{\natexlab{a}})Wang, Wang, Zhou, Ji, Gong, Zhou, Li,
  and Liu]{cosface}
Hao Wang, Yitong Wang, Zheng Zhou, Xing Ji, Dihong Gong, Jingchao Zhou, Zhifeng
  Li, and Wei Liu.
\newblock Cosface: Large margin cosine loss for deep face recognition.
\newblock In \emph{2018 {IEEE} Conference on Computer Vision and Pattern
  Recognition, {CVPR} 2018, Salt Lake City, UT, USA, June 18-22, 2018}, pages
  5265--5274. {IEEE} Computer Society, 2018{\natexlab{a}}.

\bibitem[Wang et~al.(2019)Wang, Gong, Li, and Liu]{DAL}
Hao Wang, Dihong Gong, Zhifeng Li, and Wei Liu.
\newblock Decorrelated adversarial learning for age-invariant face recognition.
\newblock In \emph{{IEEE} Conference on Computer Vision and Pattern
  Recognition, {CVPR} 2019, Long Beach, CA, USA, June 16-20, 2019}. Computer
  Vision Foundation / {IEEE}, 2019.

\bibitem[Wang et~al.(2020)Wang, Zhang, Wang, Fu, Shi, and Mei]{MV-softmax}
Xiaobo Wang, Shifeng Zhang, Shuo Wang, Tianyu Fu, Hailin Shi, and Tao Mei.
\newblock Mis-classified vector guided softmax loss for face recognition.
\newblock In \emph{The Thirty-Fourth {AAAI} Conference on Artificial
  Intelligence, {AAAI} 2020, The Thirty-Second Innovative Applications of
  Artificial Intelligence Conference, {IAAI} 2020, The Tenth {AAAI} Symposium
  on Educational Advances in Artificial Intelligence, {EAAI} 2020, New York,
  NY, USA, February 7-12, 2020}. {AAAI} Press, 2020.

\bibitem[Wang et~al.(2018{\natexlab{b}})Wang, Gong, Zhou, Ji, Wang, Li, Liu,
  and Zhang]{OECNN}
Yitong Wang, Dihong Gong, Zheng Zhou, Xing Ji, Hao Wang, Zhifeng Li, Wei Liu,
  and Tong Zhang.
\newblock Orthogonal deep features decomposition for age-invariant face
  recognition.
\newblock In \emph{Computer Vision - {ECCV} 2018 - 15th European Conference,
  Munich, Germany, September 8-14, 2018, Proceedings, Part {XV}},
  2018{\natexlab{b}}.

\bibitem[Yi et~al.(2014)Yi, Lei, Liao, and Li]{casia}
Dong Yi, Zhen Lei, S.~Liao, and S.~Li.
\newblock Learning face representation from scratch.
\newblock \emph{ArXiv}, abs/1411.7923, 2014.

\bibitem[Zhao et~al.(2019)Zhao, Cheng, Cheng, Yang, Zhao, Li, Liu, Yan, and
  Feng]{look-across-elapse}
Jian Zhao, Yu~Cheng, Yi~Cheng, Yang Yang, Fang Zhao, Jianshu Li, Hengzhu Liu,
  Shuicheng Yan, and Jiashi Feng.
\newblock Look across elapse: Disentangled representation learning and
  photorealistic cross-age face synthesis for age-invariant face recognition.
\newblock In \emph{Proceedings of the AAAI conference on artificial
  intelligence}, 2019.

\bibitem[{Zheng} et~al.(2017){Zheng}, {Deng}, and {Hu}]{AECNN}
T.~{Zheng}, W.~{Deng}, and J.~{Hu}.
\newblock Age estimation guided convolutional neural network for age-invariant
  face recognition.
\newblock In \emph{2017 IEEE Conference on Computer Vision and Pattern
  Recognition Workshops (CVPRW)}, 2017.

\end{thebibliography}
\clearpage
\appendix

\noindent \textbf{\huge{\textcolor{blue_js}{Supplementary Materials}}}

\vspace{0.5cm}
This supplementary presents 1) how we build the test sets with child-adult pairs, 2) further implementation details, and 3) comparison with re-weighting and oversampling methods. Our codes and test sets are available at \url{https://github.com/leebebeto/Inter-Prototype}.

\vspace{-0.5cm}
\section{Test sets with child-adult pairs}
\label{sec:test-set}
\vspace{-0.2cm}
Since our work focuses on face recognition with child-adult pairs, we build new test sets using the existing cross-age test sets including AgeDB~\cite{agedb} and FG-NET~\cite{fgnet}.
As mentioned in the main paper, these cross-age test sets do not necessarily include child images in each pair since they did not focus on image pairs with \emph{child-adult pairs}. 
Due to this fact, we intentionally include the child images and make pairs with a certain age gap.
For the face verification, we pair the child and the adult images of the same identity (\ie positive pairs) and different identities (\ie negative pairs) with a certain age gap.
Then, we randomly sample the equal number of positive and negative pairs. 

For the face identification, we select identities which have child and adult images with a certain age gap similar to the face verification.
Since face identification is the task of matching a given image (\ie probe image) to one of the candidate images(\ie gallery set), we use child images as the probe image and adult images as the gallery set.
The adult images in the gallery set need to have a certain age gap with child images of its identity and other identities.
After such pre-processing, FG-NET-C20 and FG-NET-C30 only include 38 and 22 identities for the gallery set, respectively. 
Due to the small number of identities in the gallery set, we \emph{did not} use these two test sets for the face identification. 
Table~\ref{tab:dataset} shows the test datasets used in our work.

\begin{table}[b!]
\centering
\scalebox{0.85}{
\begin{tabular}{cccc}
\hline
Test dataset                   & \# of pairs  & \# of gallery            \\ \hline \hline
 AgeDB-C20                     & 3786             &          63            \\ 
 FG-NET-C20                    & 1034             &           -            \\ 
 AgeDB-C30                     & 2778             &          62            \\ 
 FG-NET-C30                    & 372              &           -            \\ 
 LAG~\cite{lag}                & 3400             &          1005          \\ \cline{1-3}
\end{tabular}}
\vspace{0.25cm}
\caption{Test datasets with child-adult pairs. The second and third column indicates the number of pairs and the number of gallery images used for the face verification and face identification, respectively. We did not use FG-NET for face identification since the number of gallery images was insufficient.}
\label{tab:dataset}
\end{table}

\section{Further Implementation Details}
\label{sec:implement}

\begin{figure}[h!]
    \centering
    \includegraphics[width=0.8\textwidth, clip]{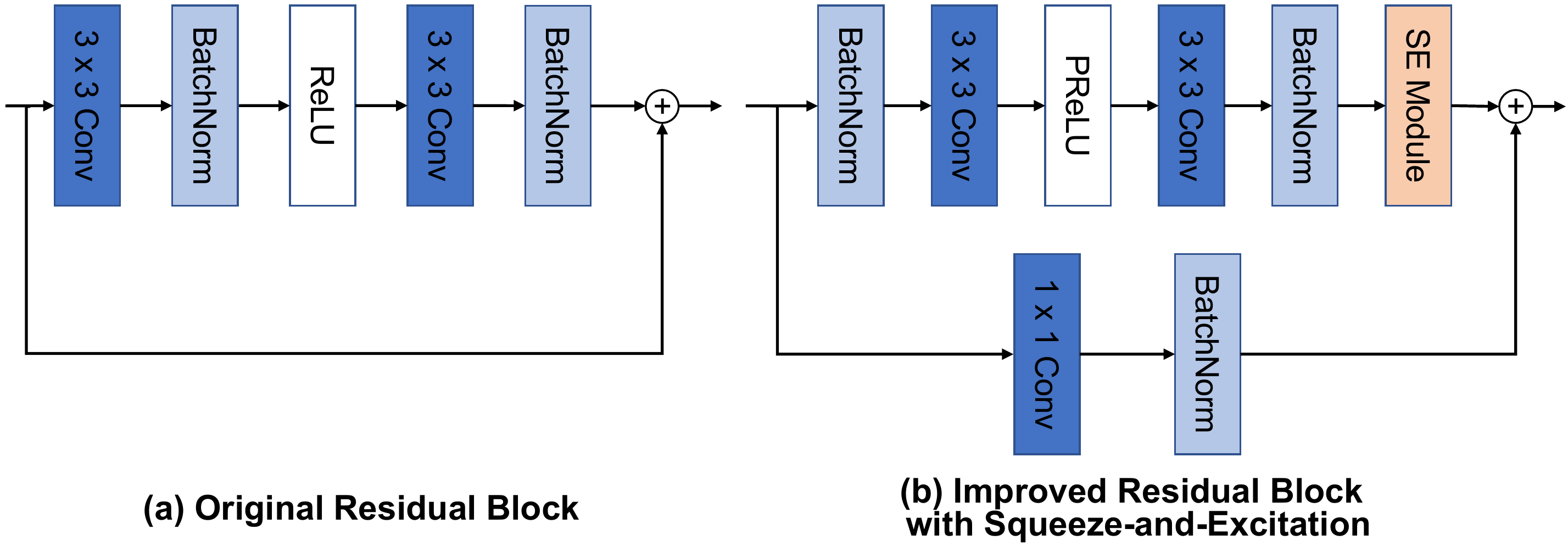}
    \vspace{-0.2cm}
    \caption{Architecture of the original Residual Block and the Improved Residual Block.}
    \label{fig:ir_se}
\end{figure}

Our backbone model for extracting features vectors from face images is based on the ResNet model~\cite{resnet}.
We adopt a variant of the ResNet used in the ArcFace paper~\cite{arcface} with a different Residual Block, called the Improved Residual Block, which has a different ordering of layers in the bottleneck layers.
As shown in Fig.~\ref{fig:ir_se}, the Improved Residual Block uses an additional convolution filter followed by Batch Normalization in the residual connection. 
The Improved Residual Block also adds a Squeeze-and-Excitation module~\cite{squeeze} before residual connections.
Note that we use the same backbone network for all experiments including our baseline models.

\vspace{-0.5cm}
\section{Extended Results on Comparison with Re-weighting/Oversampling}
\label{sec:reweight}
This section provides the comparisons between our approach and straight-forward alternatives for reducing the inter-class similarity between child images.
Those approaches are 1) re-weighting the margin-based loss function for child images with a scale of $w$, 2) assigning higher margins than the default value ($m$=0.5), and 3) oversampling child images with the ratio of $\rho$. $\rho$ indicates the ratio between the number of child images and the number of adult images in a mini-batch.
Table~\ref{tab:verification_reweight_full} compares our approach with such approaches on the test sets with child-adult pairs.
We conducted extensive experiments to find the best hyper-parameter for each approach.
Even with an extensive hyper-parameter search, none of the alternative methods outperform our proposed method.

\begin{table}[h]
\centering
\scalebox{0.8}{
\begin{tabular}{cccccc} \hline
Method                            & AgeDB-C20 & FG-NET-C20 & AgeDB-C30 & FG-NET-C30 & LAG~\cite{lag} \\ \hline
ArcFace~\cite{arcface}            & 82.65\stdv{0.31} & 80.37\stdv{1.05} & 81.91\stdv{0.55} & 79.03\stdv{0.94} & 89.80\stdv{0.25} \\ 
Re-weighting ($w$= 2, 5, 10, 50)  & 0.5 & 0.5 & 0.5 & 0.5 & 0.5 \\
Margin ($m$=0.55)                  & 82.73\stdv{0.64} & 79.75\stdv{0.93} & 81.94\stdv{0.68} & 77.96\stdv{0.47} & 89.94\stdv{0.19} \\
Margin ($m$=0.60)                  & 82.60\stdv{0.13} & 79.85\stdv{1.02} & 81.99\stdv{0.80} & 78.05\stdv{1.21} & 89.56\stdv{0.40} \\
Margin ($m$=0.65)                  & 82.83\stdv{0.09} & 80.88\stdv{0.39} & 81.88\stdv{0.63} & 77.78\stdv{1.55} & 89.90\stdv{0.11} \\
Margin ($m$=0.70)                  & 83.12\stdv{0.10} & 80.24\stdv{0.85} & 81.66\stdv{0.33} & 77.77\stdv{1.24} & 89.79\stdv{0.22} \\
Margin ($m$=0.75)                  & 82.87\stdv{0.54} & 81.04\stdv{0.93} & 82.07\stdv{0.33} & 79.30\stdv{0.71} & 89.38\stdv{0.37} \\
Margin ($m$=0.80)                  & 83.19\stdv{0.47} & 80.27\stdv{0.67} & 82.69\stdv{0.42} & 76.88\stdv{0.97} & 90.11\stdv{0.38} \\
Margin ($m$=0.85)                  & 83.01\stdv{0.25} & 80.75\stdv{0.70} & 82.02\stdv{0.70} & 78.94\stdv{0.94} & 90.13\stdv{0.54} \\
Margin ($m$=0.90)                  & 82.88\stdv{0.37} & 80.21\stdv{0.39} & 82.22\stdv{0.38} & 78.32\stdv{1.38} & 90.15\stdv{0.33} \\
Oversampling ($\rho$=0.25)        & 80.84\stdv{0.95} & 80.92\stdv{1.29} & 80.06\stdv{1.40} & 79.48\stdv{1.71} & 88.37\stdv{0.66} \\
Oversampling ($\rho$=0.5)         & 80.83\stdv{0.36} & 80.14\stdv{0.78} & 80.13\stdv{0.38} & 79.03\stdv{1.50} & 88.89\stdv{0.32} \\  
Oversampling ($\rho$=0.75)        & 80.02\stdv{1.04} & 79.14\stdv{0.15} & 79.72\stdv{0.39} & 77.24\stdv{0.95} & 88.55\stdv{0.21} \\
Oversampling ($\rho$=1.00)        & 80.06\stdv{0.97} & 79.08\stdv{1.39} & 79.22\stdv{1.01} & 78.23\stdv{2.39} & 88.23\stdv{0.54} \\\hline \hline
Ours                              & \textbf{83.55}\stdv{0.20} & \textbf{82.27}\stdv{1.03} & \textbf{82.97}\stdv{0.76} & \textbf{81.00}\stdv{1.27} & \textbf{90.49}\stdv{0.17} \\ \hline
\end{tabular}}
\vspace{0.1cm}
\caption{Comparison of our approach with re-weighting, adding high margin, and oversampling on face verification accuracy (\%).}
\vspace{-0.1cm}
\label{tab:verification_reweight_full}
\end{table}

\end{document}